\def\year{2016}\relax
\documentclass[letterpaper]{article}
\usepackage{aaai16}
\usepackage{times}
\usepackage{helvet}
\usepackage{courier}
\usepackage{amsthm}
\usepackage{color}
\usepackage{listings}
\usepackage{graphicx}
\usepackage{epstopdf}
\usepackage{epsfig}
\usepackage{amsmath}
\usepackage{enumitem}

\newtheorem{theorem}{Theorem}
\newtheorem*{verblist*}{Verb List}

\newtheorem{definition}{Definition}
\newtheorem{example}{Example}

\newtheorem{pdef}{Problem Definition}
\newcommand{\nop}[1]{}

\DeclareMathOperator*{\argmax}{arg\,max}
\DeclareMathOperator*{\argmin}{arg\,min}

\begin{document}
%
\title{Verb Pattern: A Probabilistic Semantic Representation on Verbs}


\author{ Wanyun Cui, Xiyou Zhou, Hangyu Lin, Yanghua Xiao\thanks{\small Correspondence author. This paper was supported by the National
NSFC(No.61472085, 61171132, 61033010), by National Key
Basic Research Program of China under No.2015CB358800, by
Shanghai Municipal Science and Technology Commission foundation key project under No.15JC1400900.
Seung-won Hwang was supported by Microsoft.} \\
Shanghai Key Laboratory of Data Science, School of Computer Science, Fudan University\\
wanyuncui1@gmail.com, \{ xiyouzhou13, 14302010017, shawyh\}@fudan.edu.cn
\AND
Haixun Wang\\
Facebook, USA \\
haixun@gmail.com
\And
Seung-won Hwang\\
Yonsei University\\
seungwonh@yonsei.ac.kr
\And
Wei Wang\\
Shanghai Key Laboratory of Data Science\\
School of Computer Science, Fudan Uni.\\
weiwang1@fudan.edu.cn
}



\maketitle

\begin{abstract}
Verbs are important in semantic understanding of natural language.
Traditional verb representations, such as FrameNet, PropBank, VerbNet, focus on verbs' roles. These roles are too coarse to represent verbs' semantics.
In this paper, we introduce verb patterns to represent verbs' semantics, such that each pattern corresponds to a single semantic of the verb. First we analyze the principles for verb patterns: generality and specificity. Then we propose a nonparametric model based on description length.
Experimental results prove the high effectiveness of verb patterns. We further apply verb patterns to context-aware conceptualization, to show that verb patterns are helpful in semantic-related tasks.
\end{abstract}

\section{Introduction}

Verb is crucial in sentence understanding~\cite{ferreira1990use,wu1994verbs}. A major issue of verb understanding is polysemy~\cite{rappaport1998building}, which means that {\it a verb has different semantics or senses when collocating with different objects}. In this paper, we only focus on verbs that collocate with objects. As illustrated in Example~\ref{example:polysemy}, most verbs are polysemous. Hence, a good semantic representation of verbs should be aware of their polysemy. 

\begin{example}[Verb Polysemy] {\tt eat} has the following senses:
\label{example:polysemy}
\begin{itemize}
\item a. Put food in mouth, chew it and swallow it, such as {\tt eat apple} and {\tt eat hot dog}.
\item b. Have a meal, such as {\tt eat breakfast}, {\tt eat lunch}, and {\tt eat dinner}. 
\item c. Idioms, such as {\tt eat humble pie}, which means admitting that you are wrong.
\end{itemize}
\end{example}


Many typical verb representations, including FrameNet~\cite{baker1998berkeley}, PropBank~\cite{kingsbury2002treebank}, and VerbNet~\cite{schuler2005verbnet}, 
describe verbs' semantic roles (e.g. ingestor and ingestibles for ``eat''). However, semantic roles in general are too coarse to differentiate a verb's fine-grained semantics. A verb in different phrases can have different semantics but similar roles. In Example~\ref{example:polysemy}, both ``eat''s in ``eat breakfast'' and ``eat apple'' have ingestor. But they have different semantics.

The unawareness of verbs' polysemy makes traditional verb representations unable to fully understand the verb in some applications. In sentence {\tt I like eating pitaya}, people directly know ``pitaya'' is probably one kind of food since eating a food is the most fundamental semantic of ``eat''. This enables context-aware conceptualization of pitaya to food concept. But by only knowing pitaya's role is the ``ingestibles'', traditional representations cannot tell if pitaya is a food or a meal.



{\bf Verb Patterns}
We argue that verb patterns (available at http://kw.fudan.edu.cn/verb) can be used to represent more fine-grained semantics of a verb. We design verb patterns based on two word collocations principles proposed in corpus linguistics~\cite{sinclair1991corpus}: {\it  idiom principle} and {\it open-choice principle}. 
Following the principles, we designed two types of verb patterns.

\begin{itemize}
\item \textbf{Conceptualized patterns} According to open-choice principle, a verb can collocate with any objects. Objects have certain concepts, which can be used for semantic representation and sense disambiguation~\cite{wu2012probase}. This motivates us to {\it use the objects' concepts to represent the semantics of verbs}. In Example~\ref{example:polysemy}, {\tt eat breakfast} and {\tt eat lunch} have similar semantics because both objects have concept {\tt meal}. Thus, we replace the object in the phrase with its concept to form a conceptualized pattern {\it verb \$$_C$concept} (e.g. eat \$$_C$food). Each verb phrase in open-choice principle is assigned to one conceptualized pattern according to the object's concept.   
\item \textbf{Idiom patterns} According to idiom principle, some verb phrases have specific meanings unrelated to the object's concept. We add \$$_I$ before the object to denote the idiom pattern ( i.e. {\it verb \$$_I$object}). 
\end{itemize}

According to the above definitions, we use verb patterns to represent the verb's semantics. Phrases assigned to the same pattern have similar semantics, while those assigned to different patterns have different semantics. By verb patterns, we know the ``pitaya'' in {\tt I like eating pitaya} is a food by mapping ``eat pitaya'' to ``eat \$$_C$food''. On the other hand, idiom patterns specify which phrases should not be conceptualized.
We list verb phrases from Example~\ref{example:polysemy} and their verb patterns in Table~\ref{tab:pattern}. And we will show how context-aware conceptualization benefits from our verb patterns in the application section.

\begin{table}[!htb]
\small
\begin{center}
\begin{tabular}{  l | l | l}
\hline
Verb Phrase & Verb Pattern & Type\\ \hline
  \hline
  eat apple & eat \$$_C$food & conceptualized \\ \hline
  eat hot dog & eat \$$_C$food & conceptualized  \\ \hline
  eat breakfast  & eat \$$_C$meal & conceptualized \\ \hline
  eat lunch &  eat \$$_C$meal & conceptualized \\ \hline
  eat dinner &  eat \$$_C$meal & conceptualized \\ \hline
  eat humble pie  & eat \$$_I$humble\ pie & idiom \\ \hline
\end{tabular}
\vspace{-0.2cm}
\caption{Verb phrases and their patterns}
\label{tab:pattern}
\end{center}
\vspace{-0.5cm}
\end{table}

Thus, our problem is {\it how to generate conceptualized patterns and idiom patterns for verbs}.
We use two public data sets for this purpose: Google Syntactic N-Grams (http://commondatastorage.googleapis.com/books/syntactic -ngrams/index.html) and Probase~\cite{wu2012probase}.
Google Syntactic N-grams contains millions of verb phrases, which allows us to mine
rich patterns for verbs. Probase contains rich concepts for instances, which enables the conceptualization for objects.
Thus, our problem is given a verb $v$ and a set of its phrases, generating a set of patterns (either conceptualized patterns or idiom patterns) for $v$.
However, the pattern generation for verbs is non-trivial. In general, the most critical challenge we face is the trade-off between {\it generality} and {\it specificity} of the generated patterns, as explained below.

\subsection{Trade-off between Generality and Specificity}
\label{sec:principle}


We try to answer the question: ``{\it what are good verb patterns to summarize a set of verb phrases}?'' This is hard because in general we have multiple candidate verb patterns. Intuitively, good verb patterns should be aware of the {\it generality} and {\it specificity}.


\begin{figure}[h]
\centering
\includegraphics[scale=0.45]{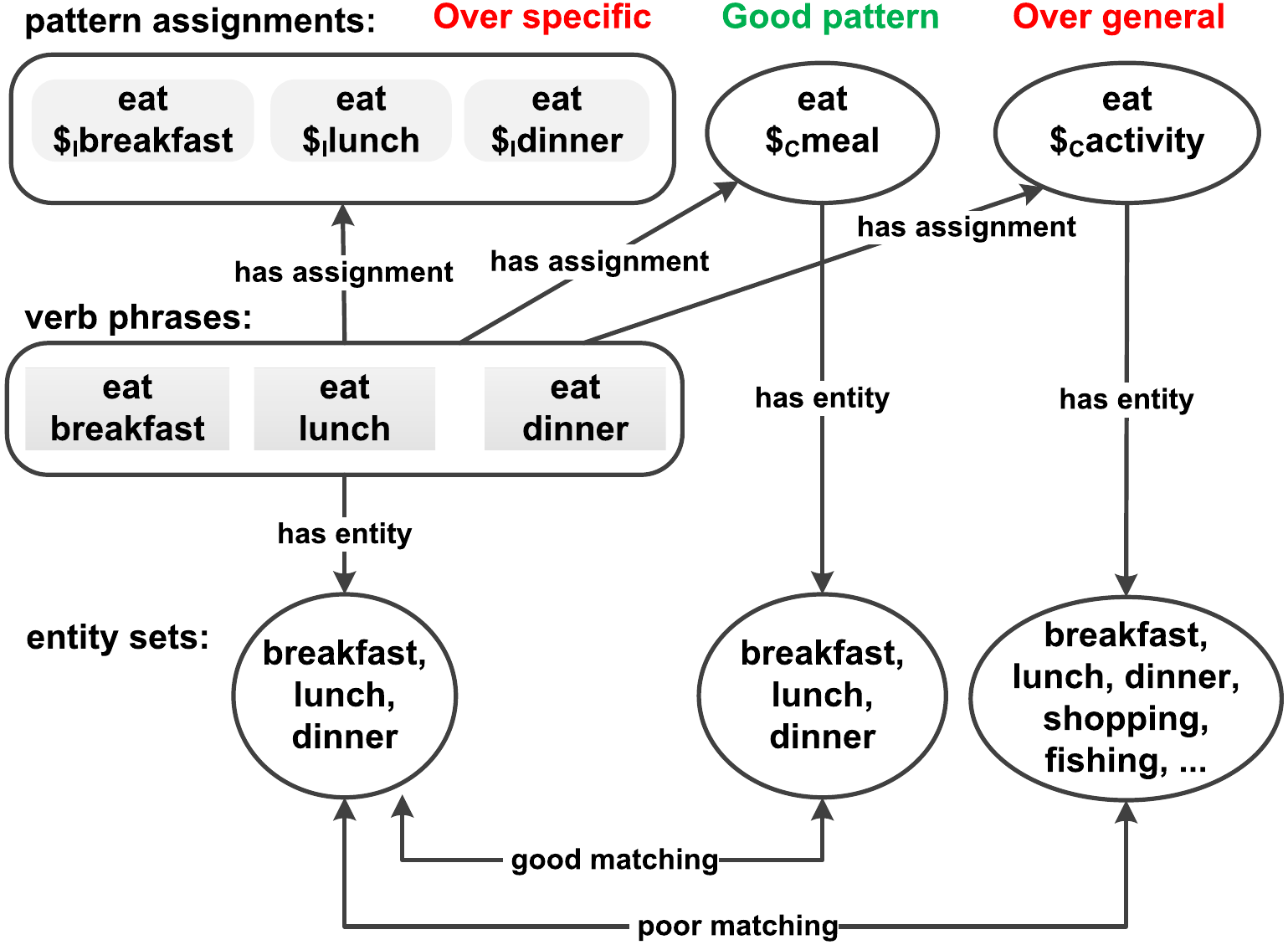}
\caption{Examples of Pattern Assignments}
\label{fig:patternassignment} 
\vspace{-0.2cm}
\end{figure}



{\bf Generality}
In general, we hope to use fewer patterns to represent the verbs' semantics. Otherwise, the extracted patterns will be trivial. Consider one extreme case where all phrases are considered as idiom phrases. Such idiom patterns obviously make no sense since idioms in general are a minority of the verb phrases.

\begin{example}
In Fig~\ref{fig:patternassignment},  ({\tt eat \$$_C$meal}) is obviously better than the three patterns ({\tt eat \$$_I$breakfast} + {\tt eat \$$_I$lunch}+ {\tt eat \$$_I$dinner}). The former case provides a more general representation. 
\end{example}




{\bf Specificity} On the other hand, we expect the generated patterns are {\it specific} enough, or the results might be trivial. As shown in Example~\ref{exa:spec}, we can generate the objects into some high-level concepts such as {\tt activity, thing}, and {\tt item}. These conceptualized patterns in general are too vague to characterize a verb's fine-grained semantic. 

\begin{example}
For phrases in Fig\ref{fig:patternassignment},  {\tt eat \$$_C$activity} is more general than {\tt eat \$$_C$meal}. As a result,  some wrong verb phrases such as {\tt eat shopping} or  {\tt each fishing} can be recognized as a valid instance of phrases for {\tt eat}. Instead, {\tt eat \$$_C$meal} has good specificity. This is because {\tt breakfast}, {\tt lunch}, {\tt dinner} are three typical instances of  {\tt meal}, and {\tt meal} has few other instances.
\label{exa:spec}
\end{example}

{\bf Contributions}
Generality and specificity obviously contradict to each other. How to find a good trade-off between them is the main challenge in this paper. We will use minimum description length (MDL) as the basic framework to reconcile the two objectives.
More specifically, our contribution in this paper can be summarized as follows:
\begin{itemize}
\item We proposed verb patterns, a novel semantic representations of verb. We proposed two types of verb patterns: conceptualized patterns and idiom patterns. The verb pattern is polysemy-aware so that we can use it to distinguish different verb semantics. 
\item We proposed the principles for verb pattern extraction: generality and specificity. We show that the trade-off between them is the main challenge of pattern generation. We further proposed an unsupervised model based on minimum description length to generate verb patterns.
\item We conducted extensive experiments. The results verify the effectiveness of our model and algorithm. We presented the applications of verb patterns in context-aware conceptualization. The application justifies the effectiveness of verb patterns to represent verb semantics.
\end{itemize}


\section{Problem Model}
\label{sec:problem}


In this section, we define the problem of extracting patterns for verb phrases. The goal of pattern extraction is to compute: (1) the pattern for each verb phrase; (2) the pattern distribution for each verb.
Next, we first give some preliminary definitions. Then we formalize our problem based on minimum description length.
The patterns of different verbs are independent from each other. Hence, we only need to focus on each individual verb and its phrases.
In the following text, we discuss our solution with respect to a given verb.


\subsection{Preliminary Definitions}
\label{sec:model:definition}
First, we formalize the definition of {\it verb phrase}, {\it verb pattern}, and {\it pattern assignment}. A verb phrase $p$ is in the form of {\tt verb + object} (e.g. ``eat apple''). We denote the object in $p$ as $o_p$. A verb pattern is either an idiom pattern or a conceptualized pattern. \textbf{Idiom Pattern} is in the form of {\tt verb \$$_I$object} (e.g. eat \$$_I$humble pie). \textbf{Conceptualized Pattern} is in the form of {\tt verb \$$_C$concept} (e.g. eat \$$_C$meal). We denote the concept in a conceptualized pattern $a$ as $c_a$.

\begin{definition}[Pattern Assignment]
A pattern assignment is a function $f: P \rightarrow A$ that maps an arbitrary phrase $p$ to its pattern $a$. $f(p)=a$ means the pattern of $p$ is $a$. The assignment has two constraints:
\begin{itemize}
\item For an idiom pattern {\tt verb \$$_I$object}, only phrase {\tt verb object} can map to it.
\item For a conceptualized pattern {\tt verb \$$_C$concept}, a phrase {\tt verb object} can map to it only if the {\tt object} belongs to the {\tt concept} in Probase~\cite{wu2012probase}.
\end{itemize}
\end{definition}

An example of verb phrases, verb patterns, and a valid pattern assignment is shown in Table~\ref{tab:pattern}.


We assume the phrase distribution is known (in our experiments, such distribution is derived from Google Syntactic Ngram). So the goal of this paper is to find $f$. With $f$, we can easily compute the pattern distribution $P(A)$ by:
\begin{equation}
\small
P(a)= \sum_p P(a|p) P(p) = \sum_{p \text{ s.t. } f(p)=a} P(p)
\label{eqn:patterndistribution}
\end{equation}, where $P(p)$ is the probability to observe phrase $p$ in all phrases of the verb of interest.
Note that the second equation holds due to the obvious fact that $P(a|p)=1$ when $f(p)=a$.
$P(p)$ can be directly estimated as the ratio of $p$'s frequency as in Eq~\ref{eq:pp}.

\subsection{Model}
\label{sec:model}
Next, we formalize our model based on minimum description length.
We first discuss our intuition to use Minimum Description Length (MDL)~\cite{barron1998minimum}. MDL is based on the idea of data compression. Verb patterns can be regarded as a compressed representation of verb phrases. Intuitively, if the pattern assignment provides a compact description of phrases, it captures the underlying verb semantics well.


Given verb phrases, we seek for the best assignment function $f$ that minimizes the code length of phrases.
Let $L(f)$ be the code length derived by $f$. The problem of verb pattern assignment thus can be formalized as below:
\begin{pdef}[Pattern Assignment]
\label{pdef:problem}
Given the phrase distribution $P(P)$, find the pattern assignment $f$, such that $L(f)$ is minimized:
\begin{equation}
\argmin_{f} L(f)
\end{equation}
\end{pdef}
We use a two-part encoding schema to encode each phrase.
For each phrase $p$, we need to encode its pattern $f(p)$ (let the code length be $l(p, f)$) as well as the $p$ itself given $f(p)$ (let the code length be $r(p, f)$).
Thus, we have
\begin{equation}
\small
L(f)=\sum_p P(p)L(p)=\sum_p P(p)[l(p, f)+r(p, f)]
\end{equation} Here $L(p)$ is the code length of $p$ and consists of $l(p, f)$ and $r(p, f)$.


{\bf $l(p,f)$: Code Length for Patterns} To encode $p$'s pattern $f(p)$, we need:
\begin{equation}
l(p,f)=-\log P(f(p))
\end{equation}
bits, where $P(f(p))$ is computed by Eq~\ref{eqn:patterndistribution}.
%

\nop{
For a given assignment $f$, $l(f(p))$ is independent from $\theta P(p)\log P_{\mathcal{T}}(p|f(p))$. So $L(f)$ is minimized only if $-\sum_p P(p)l(f(p))$ is also minimized. Since

\begin{equation}
\label{eqn:5}
-\sum_p P(p)l(f(p))=-\sum_a P(a)l(a)
\end{equation}
, using Kraft inequality and information inequality, we have:
\begin{equation}
\label{eqn:6}
L_{l}(p)= l(f(p))=-\log P(f(p))
\end{equation}

\nop{
Due to Kraft inequality and information inequality, when the sum of code length for all patterns is minimized, the code length of pattern $a$ is $-\log P(a)$. So the left part has code length:
\begin{equation}
L_{l}(p)= -\log P(f(p))
\end{equation}
}

We will describe how to compute $l(f(p))$ later. By decoding the left code, we already know phrase $p$'s pattern $f(p)$. 
}

{\bf $r(p,f)$: Code Length for Phrase given Pattern} After knowing its pattern $f(p)$, we use $P_{\mathcal{T}}(p|f(p))$, the probability of $p$ given $f(p)$ to encode $p$. $P_{\mathcal{T}}(p|f(p))$ is computed from Probase~\cite{wu2012probase} and is treated as a prior. Thus, we encode $p$ with code length $-\log P_{\mathcal{T}}(p|f(p))$.
To compute $P_{\mathcal{T}}(p|f(p))$, we consider two cases:
\begin{itemize}
\item Case 1: $f(p)$ is an idiom pattern. Since each idiom pattern has only one phrase, we have $P_{\mathcal{T}}(p|f(p))=1$.
\item Case 2: $f(p)$ is a conceptualized pattern. In this case, we only need to encode the object $o_p$ given the concept in $f(p)$. We leverage $P_{\mathcal{T}}(o_p|c_{f(p)})$, the probability of object $o_p$ given concept $c_{f(p)}$ (which is given by the isA taxonomy), to encode the phrase. We will give more details about the probability computation in the experimental settings.
\end{itemize}
Thus, we have
\begin{flalign}
\begin{split}
&r(p,f)=-\log P_{\mathcal{T}}(p|f(p)) \\
&=
\begin{cases}
-\log P(o_p|c_{f(p)}) & \text{$f(p)$ is conceptualized}\\
0 & \text{$f(p)$ is idiomatic }\\
\end{cases}
\end{split}
\end{flalign}


{\bf Total Length} We sum up the code length for all phrases to get the total code length $L$ for assignment $f$:
\begin{flalign}
\small
\begin{split}
&L(f)=\sum_p [P(p)l(p,f) + \theta P(p)r(p,f)] \\
&=-\sum_p [P(p)\log P(f(p)) + \theta P(p)\log P_{\mathcal{T}}(p|f(p))]
\end{split}
\end{flalign}
Note that here we introduce the parameter $\theta$ to control the relative importance of $l(p, f)$ and $r(p, f)$.
Next, we will explain that $\theta$ actually reflects the trade-off between the generality and the specificity of the patterns.

\subsection{Rationality}
\label{sec:rationality}

Next, we elaborate the rationality of our model by showing how the model reflects principles of verb patterns (i.e. generality and specificity). For simplicity, we define $L_L(f)$ and $L_R(f)$ as below to denote the total code length for patterns and total code length for phrases themselves:
\begin{equation}
\small
\label{eqn:llf}
L_L(f)=-\sum_p P(p)\log P(f(p))
\end{equation}
\begin{equation}
\small
L_R(f)=-\sum_p P(p)\log P_{\mathcal{T}}(p|f(p))
\end{equation}


{\bf Generality} We show that by minimizing $L_L(f)$, our model can find general patterns.
Let $A$ be all the patterns that $f$ maps to and $P_a$ be the set of each phrase $p$ such that $f(p)=a, a\in A$. Due to Eq~\ref{eqn:patterndistribution} and Eq~\ref{eqn:llf}, we have:
\begin{equation}
\small
L_L(f)=-\sum_{a\in A} \sum_{p\in P_a} P(p) \log P(a)=-\sum_{a} P(a) \log P(a)
\end{equation}
So $L_L(f)$ is the {\bf entropy} of the pattern distribution.
Minimizing the entropy favors the assignment that maps phrases to fewer patterns. This satisfies the generality principle.

%
%




{\bf Specificity}  We show that by minimizing $L_R(f)$, our model finds specific patterns.
The inner part in the last equation of Eq~\ref{eqn:lr} actually is the {\bf cross entropy} between $P(P|a)$ and $P_{\mathcal{T}}(P|a)$. Thus $L_R(f)$ has a small value if $P(P|a)$ and $P_{\mathcal{T}}(P|a)$ have similar distributions.
This reflects the specificity principle.
\begin{equation}
\label{eqn:lr}
\small
\begin{split}
L_R(f)=-\sum_{a\in A} \sum_{p\in P_a} P(p)\log P_{\mathcal{T}}(p|a) \\
=-\sum_{a\in A}P(a) \sum_{p\in P_a} \frac{P(p)}{P(a)}\log P_{\mathcal{T}}(p|a)\\
=-\sum_{a\in A}P(a) \sum_{p\in P_a} P(p|a)\log P_{\mathcal{T}}(p|a)
\end{split}
\end{equation}

\section{Algorithm}
\label{sec:algorithm}
In this section, we propose an algorithm based on simulated annealing to solve Problem~\ref{pdef:problem}.
We also show how we use external knowledge to optimize the idiom patterns.
We adopted a simulated annealing (SA) algorithm to compute the best pattern assignment $f$.
The algorithm proceeds as follows.
We first pick a random assignment as the initialization (initial temperature). Then, we generate a new assignment and evaluate it.
If it is a better assignment, we replace the previous assignment with it; otherwise we accept it with a certain probability (temperature reduction).
The generation and replacement step are repeated until no change occurs in the last $\beta$ iterations (termination condition).

\subsubsection{Candidate Assignment Generation}
Clearly, the candidate generation is critical for the effectiveness and efficiency of the procedure.
Next, we first present a straightforward candidate assignment generation approach. Then, we present an improved solution, which is aware of the typicality of the candidate patterns.

{\bf A Straightforward Generation} The basic unit of $f$ is a single pattern assignment for a phrase. A straightforward approsch is randomly picking a phrase $p$ and assigning it to a random new pattern $a$. To generate a valid pattern (by Definition 1), we need to ensure either (1) $a$ is the idiom pattern of $p$; or (2) $a$ is a conceptualized pattern and $c_a$ is a hypernym of $o_p$.
However, this approach is inefficient since it is slow to reach the optimum state. For a verb, suppose there are $n$ phrases and each of them has $k$ candidate patterns on average. The minimum number of iterations to reach the optimized assignment is $\frac{kn}{2}$ on average, which is unacceptable on big corpus.

{\bf Typicality-aware Generation}
We noticed that for a certain phrase, some patten is better than others due to their high typicality.
We illustrate this in Example~\ref{exa:pg}. This motivates us to assign phrases to patterns with higher typicality.

  \nop{assignment of phrases is highly related to the pattern set we used. When an idiom pattern is used, we surely assign its corresponding phrase to it. When a conceptualized pattern is used, all typical phrases of the pattern should be assigned.
}
\begin{example}
Consider {\tt eat breakfast, eat lunch}. {\tt eat \$$_C$meal} is obviously better than {\tt eat \$$_C$activity}.
Since it is more likely for a real human to think up with {\tt eat \$$_C$meal} than {\tt eat \$$_C$activity} when he/she sees the phrases.
In other words,  {\tt eat \$$_C$meal} is more typical than {\tt eat \$$_C$activity}.
\label{exa:pg}
\end{example}


More formally, for a certain phrase $p$, we define $t(p, a)$ to quantify the typicality of pattern $a$ with respect to $p$.
If $a$ is an idiom pattern, $t(p, a)$ is set to a constant $\gamma$. If $a$ is a conceptualized pattern, we use the typicality of object $o_p$ with respect to concept $c_a$ to define $t(p,a)$, where $c_a$ is the concept in pattern $a$.
Specifically, we have
\begin{equation}
\small
t(p,a)=
\begin{cases}
\gamma & a \text{ is idiomatic} \\
P_{\mathcal{T}}(o_p|c_a)P_{\mathcal{T}}(c_a|o_p) & a \text{ is conceptualized}
\end{cases}
\end{equation}
, where $P_{\mathcal{T}}(o_p|c_a)$ and $P_{\mathcal{T}}(c_a|o_p)$ can be derived from Probase~\cite{wu2012probase} by Eq~\ref{eq:ec}. That is, we consider both the probability from $c_a$ to $o_p$, and that from $o_p$ to $c_a$, to capture their mutual influence.

\subsubsection{Procedure}
Now we are ready to present the detailed procedure of our solution:
\begin{enumerate}[leftmargin=0.5cm]
\item Initialize $f^{(0)}$ by assigning each $p$ to its idiom pattern.
\item \label{step:2} Randomly select a new pattern $a$. For each $p$,
\begin{equation}
\small
f^{(i+1)}(p)=
\begin{cases}
a& t(p,a)>t(p,f^{(i)}(p)) \\
f^{(i)}(p) & otherwise \\
\end{cases}
\label{eqn:anew}
\end{equation}, where $f^{(i)}$ is the assignment in the $i$-th iteration.
\item \label{step:3} Accept $f^{(i+1)}$ with probability:{
\small
\begin{equation}
p=
\begin{cases}
1& L(f^{(i+1)})<L(f^{(i)})\\
e^{(L(f^{(i)})-L(f^{(i+1)}))/S^A} & L(f^{(i+1)}) \ge L(f^{(i)}) \\
\end{cases}
\end{equation}
}, where $L(f^{(i+1)})$ is the description length for $f^{(i+1)}$, $S$ is the number of steps performed in SA, and $A$ is a constant to control the speed of cooling process.
\item Repeat Step~\ref{step:2} and Step~\ref{step:3}, until there is no change in the last $\beta$ iterations.
\end{enumerate}

Step 2 and Step 3 distinguish our algorithm from the generic SA based solution.
In Step 2, for each randomly selected pattern $a$, we compute its typicality. If its typicality is larger than that of the currently assigned pattern, we assign the phrase to $a$.
In Step 3,  we accept the new assignment if its description length is smaller than that in the last round. Otherwise,
we accept it with a probability proportional to the exponential of $(L(f^{(i)})-L(f^{(i+1)}))/S^A$. The rationality is obvious: the larger deviation of $L(f^{(i+1)})$ from $L(f^{(i)})$, the less probable $f^{(i+1)}$ is accepted.


{\bf Complexity Analysis} Suppose there are $n$ phrases. In each iteration, we randomly select a pattern, then we compute the typicality of the pattern for all the $n$ phrases, which costs $O(n)$ time. Next, we compute the description length for $f^{(i+1)}$ by summing up all $n$ phrases' code lengths. This step also costs $O(n)$ time. Suppose our algorithm terminates after $S$ iterations. The entire complexity thus is $O(Sn)$. 

{\bf Incorporating Prior Knowledge of Idioms}
We notice that many verb idioms can be directly found from external dictionaries.
If a verb phrase can be judged as an idiom from dictionaries, it should be directly mapped to its corresponding idiom pattern. Specifically, we first crawled 2868 idiom verb phrases from an online dictionary. Then, in Step 2, when $p$ is one of such idiom phrases, we exclude it from the assignment update procedure.


\section{Experiments}
\label{exp:experiments}


\subsection{Settings}

\textbf{Verb Phrase Data} The pattern assignment uses the phrase distribution $P(p)$. To do this, we use the ``English All'' dataset in Google Syntactic N-Grams. The dataset contains counted syntactic ngrams extracted from the English portion of the Google Books corpus. It contains 22,230 different verbs (without stemming), and 147,056 verb phrases. 
For a fixed verb, we compute the probability of phrase $p$ by:
\begin{equation}
\small
P(p)=\frac{n(p)}{\sum_{p_i} n(p_i)}
\label{eq:pp}
\end{equation}
, where $n(p)$ is the frequency of $p$ in the corpus, and the denominator sums over all phrases of this verb.

\textbf{IsA Relationship} We use Probase to compute the probability of an entity given a concept $P_{\mathcal{T}}(e|c)$, as well as the probability of the concept given an entity $P_{\mathcal{T}}(c|e)$:
\begin{equation}
\small
P_{\mathcal{T}}(e|c)=\frac{n(e,c)}{\sum_{e_i} n(e_i,c)} \ \ \ \ \ \ \ \  P_{\mathcal{T}}(c|e)=\frac{n(e,c)}{\sum_{c_i} n(e,c_i)}
\label{eq:ec}
\end{equation}
,where $n(e, c)$ is the frequency that $c$ and $e$ co-occur in Probase.

\textbf{Test data}
We use two data sets to show our solution can achieve consistent effectiveness on both short text and long text.
The short text data set contains 1.6 millions of \textbf{tweets} from Twitter~\cite{go2009twitter}.
The long text data set contains 21,578 \textbf{news articles} from Reuters~\cite{apte1994automated}.


\subsection{Statistics of Verb Patterns}


Now we give an overview of our extracted verb patterns. For all 22,230 verbs, we report the statistics for the top 100 verbs of the highest frequency. After filtering noisy phrases with $n(p)<5$, each verb has 171 distinct phrases and 97.2 distinct patterns on average. 53\% phrases have conceptualized patterns. 47\% phrases have idiom patterns.
In Table~\ref{tab:patterncasestudy}, we list 5 typical verbs and their top patterns. The case study verified that (1) our definition of verb pattern reflects verb's polysemy; (2) most verb patterns we found are meaningful.

\begin{table}[!htb]
\small
\begin{center}
\begin{tabular}{  l  | l }
 \hline
 verb: {\bf feel} & \#phrase: 1355 \\ \hline
 feel \$$_C$symptom & feel pain (27103), feel chill (4571), ... \\ \hline
 feel \$$_C$emotion & feel love (5885), feel fear (5844), ... \\ \hline
\hline
 verb: {\bf eat} & \#phrase: 1258 \\ \hline
 eat \$$_C$meal & eat dinner (37660), eat lunch (22695), ... \\ \hline
 eat \$$_C$food & eat bread (29633), eat meat (29297), ... \\ \hline
 \hline
 verb: {\bf beat} & \#phrase: 681 \\ \hline
 beat \$$_I$retreat & beat a retreat (11003) \\ \hline
 beat \$$_C$instrument & beat drum (4480), beat gong (223), ...  \\ \hline
 \hline
 verb: {\bf ride} & \#phrase: 585 \\ \hline
 ride \$$_C$vehicle & ride bicycle (4593), ride bike (3862), ... \\ \hline
 ride \$$_C$animal & ride horse (18993), ride pony (1238), ... \\ \hline
 \hline
 verb: {\bf kick} & \#phrase: 470 \\ \hline
 kick \$$_I$ass & kick ass (10861) \\ \hline
 kick \$$_C$body part & kick leg (703), kick feet (336), ... \\ \hline
\end{tabular}
\caption{Some extracted patterns. The number in brackets is the phrase's frequency in Google Syntactic N-Gram. $\#phrase$ means the number of distinct phrases of the verb. }
\label{tab:patterncasestudy}
\end{center}
\vspace{-0.5cm}
\end{table}


\subsection{Effectiveness}
To evaluate the effectiveness of our pattern summarization approach, we report two metrics: (1) ($coverage$) how much of the verb phrases in natural language our solution can find corresponding patterns (2) ($precision$) how much of the phrases and their corresponding patterns are correctly matched? 
We compute the two metrics by:
\begin{equation}
\small
coverage=\frac{n\_cover}{n\_all} \ \ \ \ \ \ \  precision=\frac{n\_correct}{n\_cover}
\end{equation}
,where $n\_cover$ is the number of phrases in the test data for which our solution finds corresponding patterns, $n\_all$ is the total number of phrases, $n\_correct$ is the number of phrases whose corresponding patterns are correct. To evaluate $precision$, we randomly selected 100 verb phrases from the test data and ask volunteers to label the correctness of their assigned patterns. We regard a phrase-pattern matching is incorrect if it's either too {\it specific} or too {\it general} (see examples in Fig~\ref{fig:patternassignment}). For comparison, we also tested two baselines for pattern summarization:
\begin{itemize}
\item \textbf{Idiomatic Baseline (IB)} We treat each verb phrase as a idiom.
\item \textbf{Conceptualized Baseline (CB)} For each phrase, we assign it to a conceptualized pattern. For object $o_p$, we choose the concept with the highest probability, i.e. $\argmax_c P(c|o_p)$, to construct the pattern.
\end{itemize}


\begin{figure}[h]
\vspace{-0.5cm}
\centering
\includegraphics[scale=0.25]{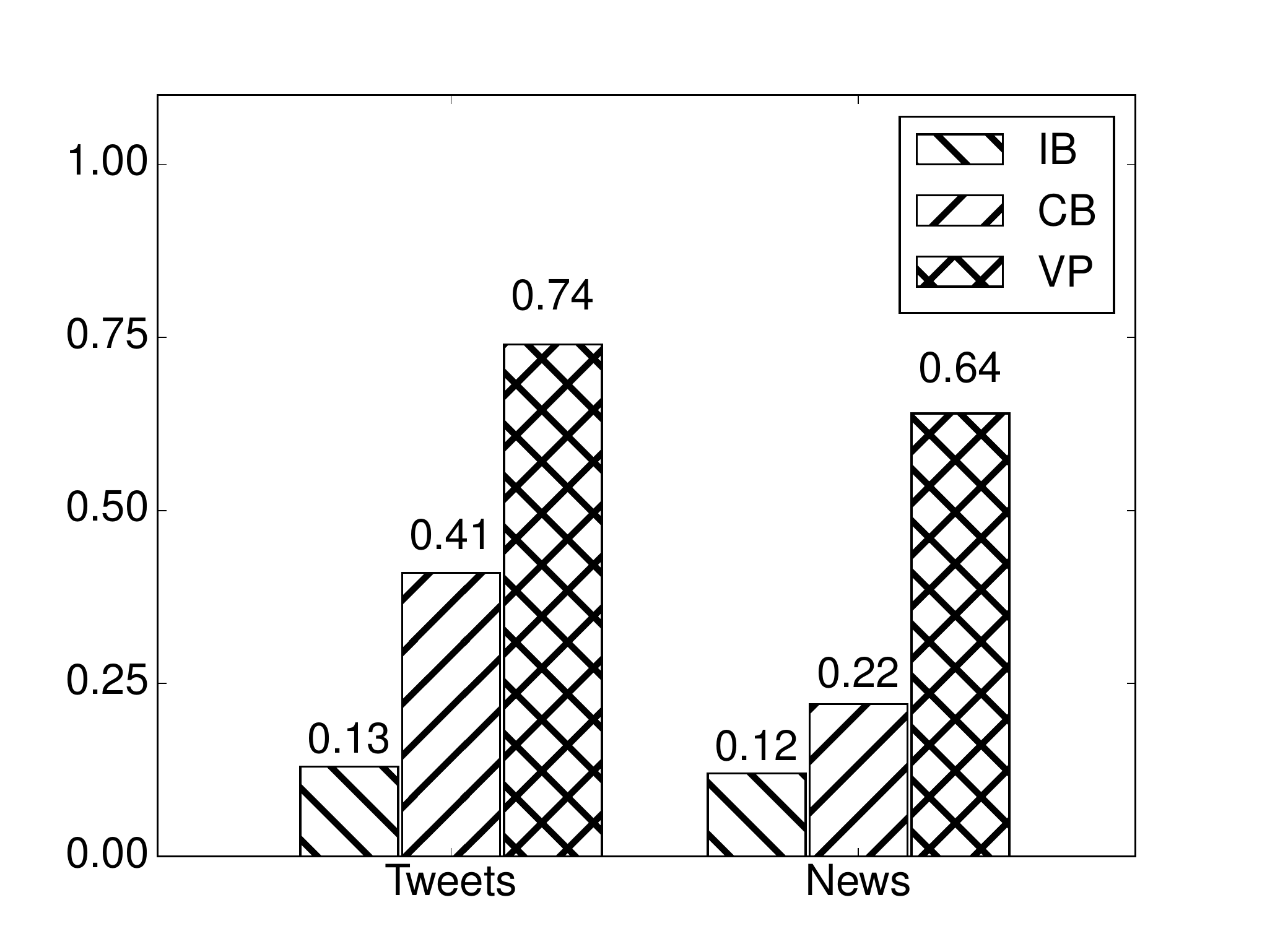}
\vspace{-0.5cm}
\caption{Precision}
\label{fig:precision}
\end{figure}

Verb patterns cover 64.3\% and 70\% verb phrases in Tweets and News, respectively. Considering the spelling errors or parsing errors in Google N-Gram data, the coverage in general is acceptable. We report the precision of the extracted verb patterns (VP) with the comparisons to baselines in Fig~\ref{fig:precision}. The results show that our approach (VP) has a significant priority over the baselines in terms of precision. The result suggests that both conceptualized patterns and idiom patterns are necessary for the semantic representation of verbs.

\section{Application: Context-Aware Conceptualization}
\label{sec:application}


As suggested in the introduction, we can use verb patterns to improve context-aware conceptualization (i.e. to extract an entity's concept while considering its context). We do this by incorporating the verb patterns into a state-of-the-art entity-based approach~\cite{song2011short}.





\textbf{Entity-based approach}
The approach conceptualizes an entity $e$ by fully employing the mentioned entities in the context.
Let $E$ be entities in the context. We denote the probability that $c$ is the concept of $e$ given the context $E$ as $P(c|e,E)$. By assuming all these entities are independent for the given concept, we compute $P(c|e,E)$ by:
\begin{equation}
\small
\label{eqn:pceE}
 P(c|e,E) \propto P(e, c)\Pi_{e_i \in E}P(e_i|c)
\end{equation}

\textbf{Our approach} We add the verb in the context as an additional feature to conceptualize $e$ when $e$ is an object of the verb. From verb patterns, we can derive $P(c|v)$, which is the probability to observe the conceptualized pattern with concept $c$ in all phrases of verb $v$. Thus, the probability of $c$ conditioned on $e$ given the context $E$ as well as verb $v$ is $P(c|e,v,E)$. Similar to Eq~\ref{eqn:pceE}, we compute it by:
{
\small
\begin{flalign}
\label{eqn:pceve}
\begin{split}
&P(c|e,v,E) = \frac{P(e,v,E|c)P(c)}{P(e,v,E)} \propto P(e, v,E|c)P(c)  \\
&= P(e|c)P(v|c)P(E|c)P(c) \\
&=P(e|c)P(c|v)P(v)\Pi_{e_i \in E} P(e_i|c) \\
&\propto P(e|c)P(c|v) \Pi_{e_i \in E} P(e_i|c)
\end{split}
\end{flalign}
}
Note that if $v+e$ is observed in Google Syntactic N-Grams, which means that we have already learned its pattern, then we can use these verb patterns to do the conceptualization. That is, if $v+e$ is mapped to a conceptualized pattern, we use the pattern's concept as the conceptualization result. If $v+e$ is an idiom pattern, we stop the conceptualization.

\begin{figure}[h]
\vspace{-0.2cm}
\centering
\includegraphics[scale=0.25]{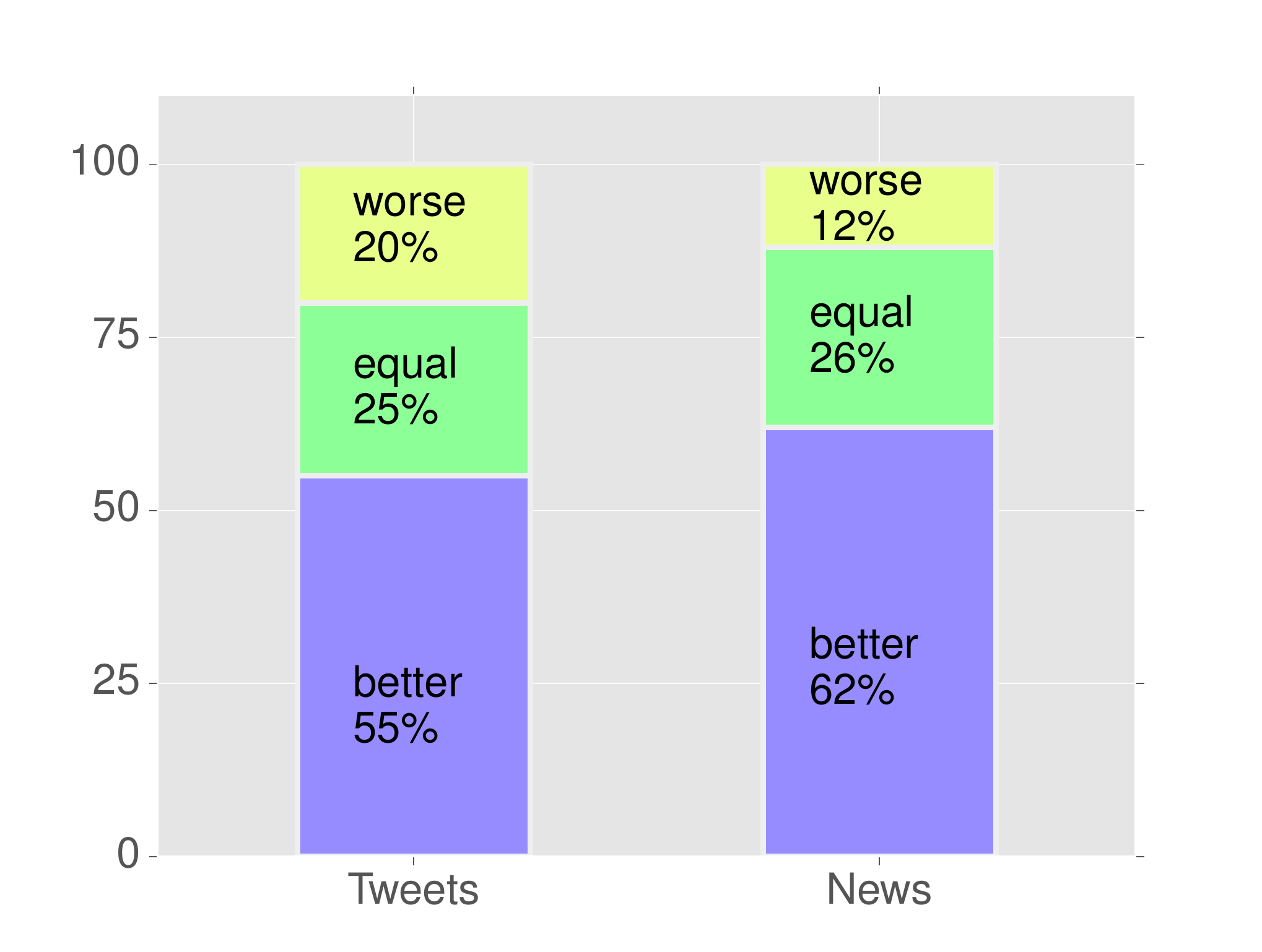}
\vspace{-0.5cm}
\caption{Conceptualization Results}
\label{fig:conceptualization} 
\vspace{-0.5cm}
\end{figure}

{\bf Settings and Results} For the two datasets used in the experimental section, we use both approaches to conceptualize objects in all verb phrases. Then, we select the concept with the highest probability as the label of the object. We randomly select 100 phrases for which the two approaches generate different labels. For each difference, we manually label if our result is {\it better} than, {\it equal} to, or {\it worse} than the competitor. Results are shown in Fig~\ref{fig:conceptualization}. On both datasets, the precisions are significantly improved after adding verb patterns.
This verifies that verb patterns are helpful in semantic understanding tasks.

\nop{
We report the results of the baseline and ours in Table~\ref{tab:conceptualization}.
\begin{table}[!htb]
\small
\begin{center}
\begin{tabular}{ |l | l | l | l |}
  \hline
    Method & worse & equal & better\\ \hline
  \hline
    Baseline~\cite{song2011short}  & - & - & - \\ \hline
    Ours II &-  & - & - \\ \hline
\end{tabular}
\caption{RI score}
\label{tab:conceptualization}
\end{center}
\end{table}
}

\nop{
how to compare
---
(environment)

In this section, we present our experimental results. The probability $P(e|c)$ is derived from *Probase*. We carried out the experiments on a PC with Intel i7 cpu@2.3Ghz and 8G memory. We implement all the programs in Python with 10027 tweets as corpus.
Specially we compare the results of our approach and baseline instead of comparing precision and recall directly because it's hard to definitely judge a conceptualization as right or wrong. However the results can be relatively better or worse. Therefore we randomly pick up 100 entities with different conceptualization results to compare our approach to naive bayes approach with three kind of labels: better equal or worse. The results are labeled manually.

1.Conceptualization
2.Comparison
}

\section{Related Work}


\textbf{Traditional Verb Representations} We compare verb patterns with traditional verb representations~\cite{palmer2009semlink}. FrameNet~\cite{baker1998berkeley} is built upon the idea that the meanings of most words can be best understood by semantic frames~\cite{fillmore1976frame}. Semantic frame is a description of a type of event, relation, or entity and the participants in it. And each semantic frame uses frame elements (FEs) to make simple annotations. PropBank~\cite{kingsbury2002treebank} uses manually labeled predicates and arguments of semantic roles, to capture the precise predicate-argument structure. The {\it predicates} here are verbs, while {\it arguments} are other roles of verb. To make PropBank more formalized, the arguments always consist of agent, patient, instrument, starting point and ending point. VerbNet~\cite{schuler2005verbnet} classifies verbs according to their syntax patterns based on Levin classes~\cite{levin1993english}. All these verb representations focus on different roles of the verb instead of the semantics of verb. While different verb semantics might have similar roles, the existing representations cannot fully characterize the verb's semantics.

\textbf{Conceptualization} One typical application of our work is context-aware conceptualization, which motivates the survey of the conceptualization.  Conceptualization determines the most appropriate concept for an entity.Traditional text retrieval based approaches use NER~\cite{tjong2003introduction} for conceptualization. But NER usually has only a few predefined coarse concepts. Wu et al. built a knowledge base with large-scale lexical information to provide richer IsA relations~\cite{wu2012probase}. Using IsA relations, context-aware conceptualization~\cite{kim2013context} performs better. Song et al.~\cite{song2011short} proposed a conceptualization mechanism by Naive Bayes. And Wen et al.~\cite{hua2015short} proposed a state-of-the-art model by combining co-occurrence network, IsA network and concept clusters.

\textbf{Semantic Composition} We represent verb phrases by verb patterns.
while semantic composition works aim to represent the meaning of an arbitrary phrase as a vector or a tree. Vector-space model is widely used to represent the semantic of single word. A straightforward approach to characterize the semantic of a phrase thus is averaging the vectors over all the phrase's words~\cite{chen2015joint}. But this approach certainly ignores the syntactic relation~\cite{landauer1997solution} between words. Socher et al.~\cite{socher2011dynamic} represent the syntactic relation by a binary tree, which is fed into a recursive neural network together with the words' vectors. Recently, word2vec~\cite{mikolov2013efficient} shows its advantage in single word representation. Mikolov et al.~\cite{mikolov2013distributed} further revise it to make word2vec capable for phrase vector. In summary, none of these works uses the idiom phrases of verbs and concept of verb's object to represent the semantics of verbs. 

%
%

\section{Conclusion}

Verbs' semantics are important in text understanding. In this paper, we proposed verb patterns, which can distinguish different verb semantics. We built a model based on minimum description length to trade-off between generality and specificity of verb patterns. We also proposed a simulated annealing based algorithm to extract verb patterns. We leverage patterns' typicality to accelerate the convergence by pattern-based candidate generation. Experiments justify the high precision and coverage of our extracted patterns. We also presented a successful application of verb patterns into context-aware conceptualization.

\newpage
\bibliography{verb}
\bibliographystyle{aaai}
\end{document}